# Teleoperation of a Humanoid Robot with Motion Imitation and Legged Locomotion

Aditya Sripada, Harish Asokan, Abhishek Warrier, Arpit Kapoor, Harshit Gaur, Sridhar R

*Abstract: This work presents a teleoperated humanoid robot system that can imitate human motions, walk and turn. To capture human motions, a Microsoft Kinect Depth sensor is used. Unlike the cumbersome motion capture suits, the sensor makes the system more comfortable to interact with. The skeleton data is extracted from the Kinect which is processed to generate the robot's joint angles. Robot Operating System (ROS) is used for the communication between the various parts of the code to achieve minimal latency. Thus the robot imitates the human in real-time with negligible lag. Unlike most of the human motion imitation systems, this system is not stationary. The lower body motions of the user are captured and processed and used to make the robot walk forward, backward and to make it turn right or left thus enabling a completely dynamic teleoperated humanoid robot system.*

## I. INTRODUCTION

In recent years, significant progress has been made in various fields such as Artificial Intelligence, Machine Learning and Cognitive Science which has made a considerable impact in the field of robotics. With the rise of techniques such as deep neural nets and reinforcement learning, machines can now be programmed to do tasks that previously seemed a far-fetched reality. But these techniques are still constrained by various factors and require enormous volumes of data and computational resources to show favorable results. Thus despite all the development, there are still many potential areas of application that robots are yet to penetrate. Teleoperation Techniques prove to be much effective instead.

The main motivation behind teleoperation is that it offers a two-fold advantage. Firstly, it enables the use of robots in areas where it is too complex to program a robot to act autonomously. Secondly, the data collected from imitation based systems such as the one proposed in this work can be used to train learning systems to eventually perform the same task autonomously.

Over the years, a lot of research has been done on imitation based systems that track human motions and map them to a humanoid robot. This is because humanoid robots are extremely versatile and a breakthrough in the teleoperation of humanoids would open many new doors. But many of these systems are stationary and with minimal focus on mobility.

Thus this work proposes a teleoperation concept that in addition to upper body imitation, would enable users to walk forward or backward and also turn the humanoid robot in real time. For tracking, the Microsoft Kinect Sensor is used which with its low cost and availability of open source packages is a favourable platform. Additionally, as Kinect doesn't need users to wear any extra device or equipment, it also offers a more immersive experience. To make the system responsive and robust, it is implemented using the Robot Operating System (ROS) ecosystem.

## II. RELATED WORK

Mukherjee *et al*. [1] described three different methods to perform human upper body motion imitation on a Nao humanoid robot. The three methods implemented are direct angle mapping method, inverse kinematics using fuzzy logic and inverse kinematics using iterative Jacobian. The results obtained from each were then compared. Nguyen *et al*. [2] proposed a system capable of imitating human motions on a humanoid robot. The two major issues relating to mapping of different kinematic structures and computing humanoid body pose were solved to achieve the same. Results based on a simulation and actual execution on a Darwin-OP robot showed the validity of the system. The work done by Yokota *et al*. [3] deals with imitation of human motions on a humanoid robot using a Kinect sensor. His approach estimates the perceptible delay time for a person when a delay exists between the motion of the robot and motion with respect to themselves. Also, human impressions are similarly investigated and the motion is statistically verified. Yavsan *et al*. [4] developed an algorithm to recognize human actions and reproduce them on a robot using a Kinect 360 sensor. The two algorithms used are Extreme Learning Machines (ELMs) and the Feed Forward Neural Networks (FNNs) and

Aditya Sripada was a final year undergraduate student with the Department of Electrical and Electronics Engineering, SRM University, Chennai
(E-mail: adityassripada@gmail.com)
Harish Asokan was a final year undergraduate student with the Department of Electrical and Electronics Engineering, SRM University, Chennai
(E-mail: harishra94@gmail.com)
Abhishek Warrier is a third year undergraduate student with the Department of Computer Science and Engineering, SRM University, Chennai
(E-mail: warrier.abhishek@gmail.com)
Arpit Kapoor is a third year undergraduate student with the Department of Computer Science and Engineering, SRM University, Chennai
(E-mail: arpitkapur.dps@gmail.com)
Harshit Gaur is a third year undergraduate student with the Department of Mechanical Engineering, SRM University, Chennai
(E-mail: hgaur014@gmail.com)
Sridhar R is with the Department of Electrical and Electronics Engineering, SRM University, Chennai as an Assistant Professor
(E-mail: sridhar.r@srmuniv.ac.in)

according to comparative results, ELMs stood out in recognition performance. Avalos *et al*. [5] developed a system that allows controlling a humanoid robot using a motion capture device. The robot is capable of sending audio and images from the environment. The system was tested using a robot called NAO which was successful. The work done by Lee *et al*. [7] investigates stability issues to prevent a humanoid from falling down during the imitation of human motions. A Kinect depth sensor was used to obtain the skeleton data. The paper proposed EFS (Estimation Function for Stability) to estimate the (COG) Centre of Gravity of the robot using a simple model and was tested using the humanoid robot, KIBO. Koenemann *et al*. [8] proposed an approach to imitate whole body motions of humans with endeffectors and center of mass of the robot as the major aspects to imitate. The human data was captured using an Xsens MVN motion capture system. The approach designed, generated a stable pose for every point in time. The results were tested on the NAO humanoid robot which was able to perform complex human motions in real-time.

### III. THE HUMANOID ROBOT

This work uses a 20DOF Humanoid robot as shown in Fig.1. The specifications of the robot are presented in the Table I. The robot has two hands, two legs, a head and a torso. The robot structure is laser cut out of 1.5mm 7075 Aluminium. The robot uses a Hard Kernel Odroid XU4 Octacore Processor board as the main computer. Twenty Robotis MX28 Smart servos are used as the actuators.

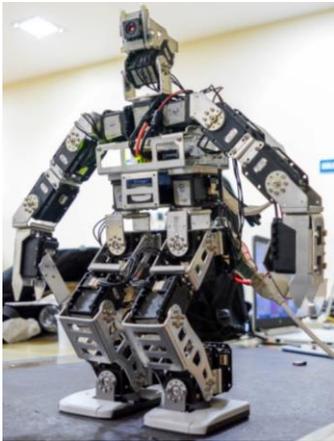

Fig.1. Humanoid Robot used for the experiment

TABLE I. Robot Specifications

| Robot | Specification |
|---|---|
| Height | 45 CM |
| Weight | 2.5 KG |
| Degrees of Freedom | 20 |
| Actuators | 20 X MX28T Dynamixel |
| Vision sensor | Logitech C170 |
| Main Computer | Odroid XU4 |
| Operating System | Linux |
| Battery | 3S 2200mAh 20C Lipo |

As Human body is arguably the most complex machine and recreating it artificially is nearly impossible, it is very hard to provide a robot with all kinds of degrees of freedom as in a human body. Human body has more than 240 degrees of freedom but due to structural constraints, the robot used for this experiment has 20 degrees of freedom. The robot's links and joints are as shown in the Fig.2. The numbers, names and the motion limits of the joints are as shown in the Table. II.

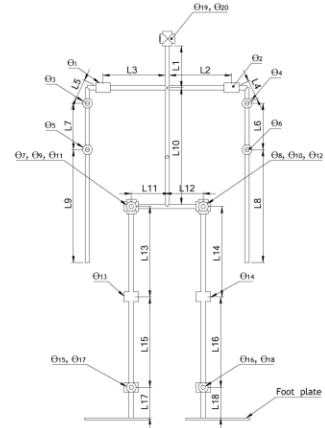

Fig2. Robot Structure and Links

TABLE II. Robot joint limits

| Joint Number | Joint Name | Axis | $\theta_{Min}$ | $\theta_{Max}$ |
|---|---|---|---|---|
| **Right Arm** | | | | |
| 1 | Shoulder Pitch | $Z_1$ | -4 $\pi$/3 | 4 $\pi$/3 |
| 3 | Shoulder Roll | $Z_2$ | - $\pi$/2 | $\pi$/2 |
| 5 | Elbow | $Z_3$ | 0 | 5$\pi$/6 |
| **Left Arm** | | | | |
| 2 | Shoulder Pitch | $Z_1$ | -4 $\pi$/3 | 4 $\pi$/3 |
| 4 | Shoulder Roll | $Z_2$ | - $\pi$/2 | $\pi$/2 |
| 6 | Elbow | $Z_3$ | -5$\pi$/6 | 0 |
| **Right Leg** | | | | |
| 7 | Hip Yaw | $Z_1$ | -5$\pi$/6 | $\pi$/4 |
| 11 | Hip Roll | $Z_2$ | 0 | $\pi$/3 |
| 9 | Hip Pitch | $Z_3$ | - $\pi$/2 | $\pi$/6 |
| 13 | Knee | $Z_4$ | 0 | 3$\pi$/4 |
| 17 | Ankle Pitch | $Z_5$ | - $\pi$/3 | $\pi$/3 |
| 15 | Ankle Roll | $Z_6$ | - $\pi$/6 | $\pi$/3 |
| **Left Leg** | | | | |
| 8 | Hip Yaw | $Z_1$ | -$\pi$/4 | 5$\pi$/6 |
| 12 | Hip Roll | $Z_2$ | - $\pi$/3 | 0 |
| 10 | Hip Pitch | $Z_3$ | - $\pi$/6 | $\pi$/2 |
| 14 | Knee | $Z_4$ | -3$\pi$/4 | 0 |
| 18 | Ankle Pitch | $Z_5$ | - $\pi$/3 | $\pi$/3 |
| 16 | Ankle Roll | $Z_6$ | - $\pi$/6 | $\pi$/3 |
| **Head** | | | | |
| 19 | Head Yaw | $Z_1$ | -5$\pi$/6 | 5$\pi$/6 |
| 20 | Head Pitch | $Z_2$ | - $\pi$/3 | $\pi$/6 |

## IV. SOFTWARE FRAMEWORK

### A. Overview

The software framework is built on the ROS Platform which provides a comprehensive set of tools and packages that allow the creation of responsive distributed systems. The main advantage of using ROS is that the code can be split into modules called nodes. Each node runs independently and communicates with other nodes based on a Publisher-Subscriber model, where messages are sent over specific channels called topics. It enables the process of human tracking to be completely independent of robot actuation making the system robust.

The system consists of three nodes: Skeleton Tracker node, Angle Computation node, and Motor Actuation node. The Skeleton_tracker node obtains skeletal data which is published over the /skeleton topic; subscribed by the Computation_node. The Motor Actuation node which runs on the Odroid-XU4 placed on the robot receives computed joint angles over the /skel_angles topic. Communication between all nodes is facilitated by a ROS Master which means that nodes running on different machines can communicate with each other. Thus, the Motor Actuation node can receive angles wirelessly from a remote machine on which the other two nodes are running.

### B. Skeletal Image Capture:

For tracking humans and capturing the skeletal image, Microsoft Kinect, which is an RGB-D camera, is used. Skeletal data from the Kinect can be obtained using either Kinect SDK or OpenNI SDK. The Kinect SDK only supports Microsoft Windows which is not completely compatible with ROS. In comparison, OpenNI is an open source framework supported across multiple platforms and is used by various ROS packages. The openni_tracker package publishes the skeletal frames as shown in the Fig.3 using tf as shown in Fig.5 which is the ROS transform library that lets users keep track of multiple frames over time. An additional package called skeleton_markers reads the transforms and publishes the data over a Skeleton message which is more convenient to access. The Skeleton message contains the position, orientation and confidence value for every tracked joint.

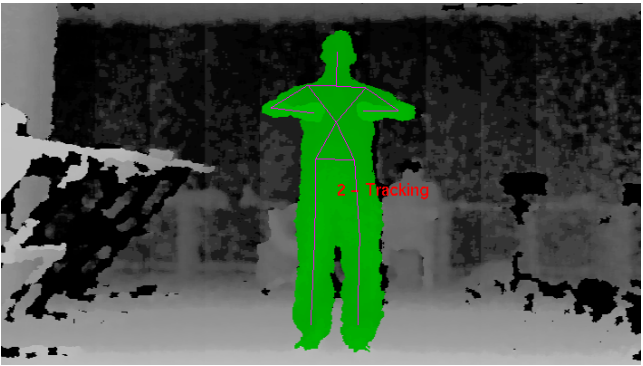

Fig.3 Skeletal tracking of human from the depth image obtained from the Kinect.

### C. Angle Computation

The skeleton data extracted from the Kinect is raw data which has coordinates in 3D Space. The skeletal image looks as shown in the Fig.4.

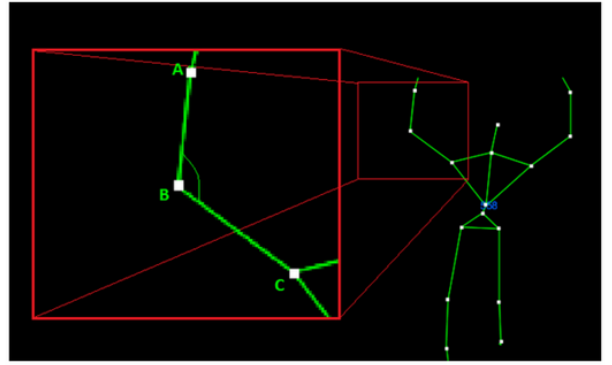

Fig.4 Skeletal Image with the right arm region zoomed in

The position values of the skeletal points from the skeleton message are converted into vectors as shown in the equations (1) and (2).

$$\overline{AB} = (A_x - B_x)\hat{\imath} + (A_y - B_y)\hat{\jmath} + (A_z - B_z)\hat{k} \quad (1)$$
$$\overline{BC} = (B_x - C_x)\hat{\imath} + (B_y - C_y)\hat{\jmath} + (B_z - C_z)\hat{k} \quad (2)$$

$$B = \cos^{-1}\left(\frac{\overline{AB}.\overline{BC}}{|AB||BC|}\right) \quad (3)$$

Where $(A_x, A_y, A_z)$, $(B_x, B_y, B_z)$ and $(C_x, C_y, C_z)$ are the locations of the skeleton points A, B and C respectively in the 3D Coordinate system. The angles between joints are computed by using the equation (3).

The Computation_node makes use of PyKDL which is a kinematics and dynamics library for python. The position and orientation of each joint are stored as KDL vectors and quaternions respectively.

### D. Motor Actuation

The Actuation Node is responsible for controlling the robot using Pypot which is a high-level python library for interfacing with Dynamixel Motors. This node is also responsible for dynamically controlling the speed of each motor which is calculated as given in equation (4).

$$\omega = \omega_0 + \omega_0 \left(\frac{|\varphi_{new} - \varphi_{prev}|}{180}\right) \quad (4)$$

Where, $\omega$ = Final Speed
$\omega_0$ = Base Speed
$\varphi_{new}$ = New Joint Angle
$\varphi_{prev}$ = Current Joint Angle

Thus, the speed of each motor is proportional to the angular displacement it needs to cover.

## V. TELEOPERATION

In the following, we present a teleoperation concept with the robot imitating the upper body human motions while being able to walk and turn with a negligible lag. The human pose is recorded by the Kinect at a rate of 20FPS. The imitation happens with an interval rather than continuously. The upper body is controlled in real time with an appreciable accuracy.

### A. Upperbody Imitation

Whenever a new set of angles is received, based on the formula given in equation (4), the speed of each motor in the upper body is set and the angles are written to the motors immediately. This restriction in the speed of the end effector ensures that the motion is smooth and at the same time assuring that the motors reach the goal position without any latency. Thus, creating a balance between the two. But a problem arises that the hand motions can interfere with the walking and turning cycles. Thus to counter this problem, the system is implemented in such a way that upper body imitation has a lower precedence compared to walking/turning. That is the upper body is only imitated when the humanoid isn't turning/walking. As walks and turns are less frequent, the humanoid is usually imitating the upper body. When a walk/turn is triggered by the user the hand motions are temporarily paused and the respective subroutine is called, and upper body imitations are resumed immediately after that. As the walk/turn subroutines last only a short duration, this approach offers a reasonable compromise between real-time teleoperation and the stability of the robot. The robot imitates the upper body motions as shown in Fig.6

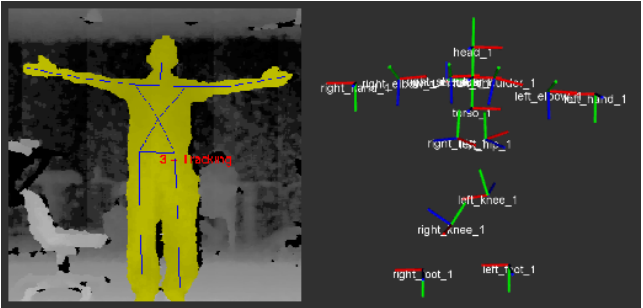

Fig.5 Data Processing for upper body imitation. Left: Skeletal image of human Right: TF frames of all body joints

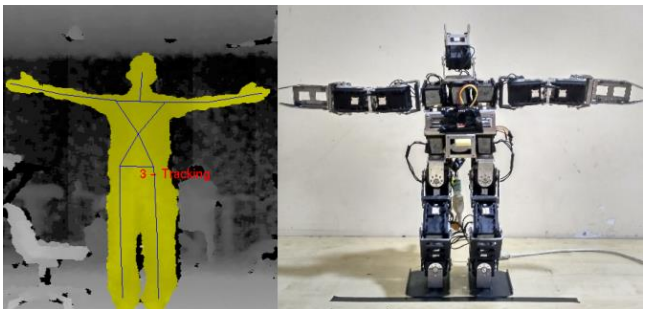

Fig.6. Upper body Imitation

### B. Walking

The Humanoid has the capability to walk both forwards and backward. Based on the human poses, the system needs to decide whether to walk forward or backward and move accordingly. To implement such a system, four values are constantly monitored which consists of the knee angle and the depth of the knee with respect to the Kinect for both legs. Whenever the operator lifts a leg, the knee angle crosses a certain threshold and that leg is marked. When the operator places the marked leg down the knee angle falls below the threshold and at that point, we calculate the difference in depth between both the knees. As there might be a slight fluctuation in the depth values, a depth threshold is also used to determine that the difference is significant and not random noise.

When both the legs are together, the robot is said to be in the initial state. When the humanoid takes a step in any direction, the robot is not in the initial state as the legs are not together anymore. The algorithm to decide which step to take is based on the depth values and the state of the humanoid and is given below in Algorithm 1. After the decision is made, the motion set for that particular gait cycle is called up and transmitted to the robot.

**Algorithm 1** Deciding which step to take

```
1    depth_diff = marked_leg.depth - unmarked_leg.depth
2    if initial_state =True then:
3            if (depth_diff ) > depth_threshold then
4                    Take back step with marked_leg
5                    Set initial_state := False
6                    Set marked_leg.state := Back
7                    Set unmarked_leg.state := Forward
8            else if  (depth_diff ) < - depth_threshold then
9                    Take forward step with marked_leg
10                   Set initial_state := False
11                   Set marked_leg.state := Forward
12                   Set unmarked_leg.state := Back
13
14           end if
15   else
16           if marked_leg.state = Forward then
17                   Take back step with marked_leg
18           else if marked_leg.state = Backward then
19                   Take forward step with marked_leg
20           end if
21
22           Set initial_state := True
23           Set marked_leg.state := Null
24           Set unmarked_leg.state := Null
25   end if
```

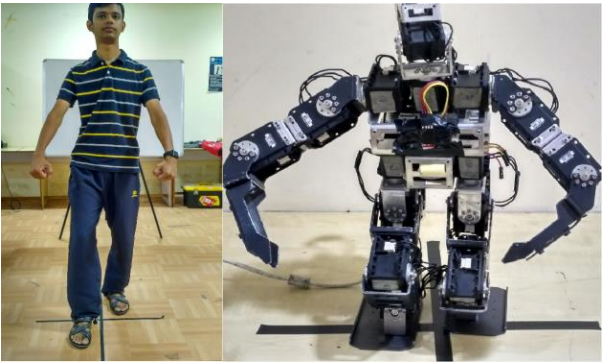
Fig.7 The robot imitating human while taking forward step.

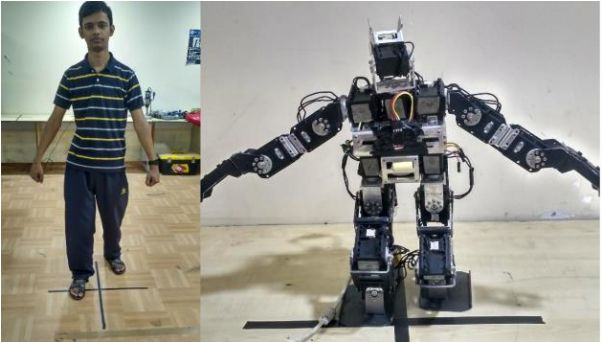
Fig.8 The robot imitating human while taking backward step.

The robot taking front and back steps based on the human pose can be seen from the Fig.7 and Fig.8 respectively.

*C. Turning*

Turning is another important aspect of the teleoperation of a humanoid robot. Just like the Walk, turning has two scenarios, right and left and the decision is made based on the human pose with respect to the Kinect. To obtain the orientation of the user, we convert the quaternion of the torso joint to Euler angles. Due to structural constraints, the humanoid can't turn like humans do in a single step, but it will be a combination of multiple steps and this can be seen from Fig.9. So when the user twists his body, the torso angle of the operator is mapped to the number of the turns that the robot needs to take in that direction.

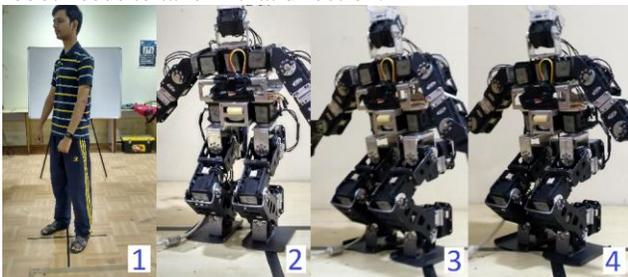
Fig.9 Snapshots of robot imitating the human while turning.
1: Human operating the robot
2, 3, 4: Multiple steps taken by robot to align its orientation with that of the human.

Youtube Video of the proposed teleoperation system:
https://www.youtube.com/watch?v=znR0XXcpGIs

## VI. CONCLUSION

This method is proposed to create a full-fledged teleoperation system with human motion imitation along with legged locomotion. The technique presented for real time teleoperation enables the user to control the robot's motions along with walking and turning which makes the system dynamic, unlike the most common static imitation techniques used for teleoperation. The robot exhibits very small, rather unnoticeable lag while imitating the human motions.